\documentclass[final]{cvpr}

\usepackage{times}
\usepackage{epsfig}
\usepackage{graphicx}
\usepackage{amsmath}
\usepackage{amssymb}

\usepackage{booktabs,tabulary,multirow,overpic,xcolor}

\usepackage{microtype}

\usepackage{listings}

\newcolumntype{x}[1]{>{\centering\arraybackslash}p{#1pt}}

\newlength\savewidth\newcommand\shline{\noalign{\global\savewidth\arrayrulewidth
  \global\arrayrulewidth 1pt}\hline\noalign{\global\arrayrulewidth\savewidth}}

\usepackage[pagebackref=true,breaklinks=true,colorlinks,bookmarks=false]{hyperref}

\def\cvprPaperID{7354} %
\def\confYear{CVPR 2021}

\usepackage{bm}
\usepackage{mathtools}

\newcommand*\vv[1]{\mathbf{#1}}

\newcommand{\RR}{\mathbb{R}}

\newcommand{\vx}{\vv{x}}

\DeclareMathOperator*{\concat}{||}
\DeclareMathOperator*{\normalize}{Norm}
\DeclareMathOperator*{\aggr}{Aggr}
\DeclareMathOperator*{\sign}{sign}

\newcommand*\mypar[1]{\vspace{0.5em} \noindent \textbf{#1} \hspace{0.5em}}

\begin{document}

\title{Binary Graph Neural Networks}%

\author{Mehdi Bahri\textsuperscript{1} \quad \quad Gaétan Bahl\textsuperscript{2,3} \quad \quad Stefanos Zafeiriou\textsuperscript{1}\\
\textsuperscript{1}Imperial College London, UK \quad \textsuperscript{2}Université Côte d’Azur - Inria \textsuperscript{3}IRT Saint-Exupéry\\
{\tt\small \{m.bahri, s.zafeiriou\}@imperial.ac.uk, gaetan.bahl@inria.fr}
}

\maketitle

\begin{abstract}
Graph Neural Networks (GNNs) have emerged as a powerful and flexible framework for representation learning on irregular data. As they generalize the operations of classical CNNs on grids to arbitrary topologies, GNNs also bring much of the implementation challenges of their Euclidean counterparts. Model size, memory footprint, and energy consumption are common concerns for many real-world applications. Network binarization allocates a single bit to parameters and activations, thus dramatically reducing the memory requirements (up to 32x compared to single-precision floating-point numbers) and maximizing the benefits of fast SIMD instructions on modern hardware for measurable speedups. However, in spite of the large body of work on binarization for classical CNNs, this area remains largely unexplored in geometric deep learning. In this paper, we present and evaluate different strategies for the binarization of graph neural networks. We show that through careful design of the models, and control of the training process, binary graph neural networks can be trained at only a moderate cost in accuracy on challenging benchmarks. In particular, we present the first dynamic graph neural network in Hamming space, able to leverage efficient $k$-NN search on binary vectors to speed-up the construction of the dynamic graph. We further verify that the binary models offer significant savings on embedded devices. Our code is publicly available on Github\footnote{\href{https://github.com/mbahri/binary_gnn}{https://github.com/mbahri/binary\_gnn}}.
\end{abstract}

\begin{figure}[t]
    \centering
    \includegraphics[width=\linewidth]{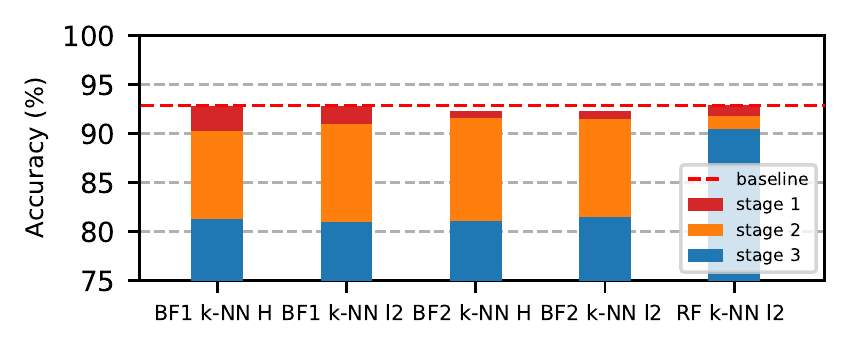}
    \includegraphics[width=\linewidth]{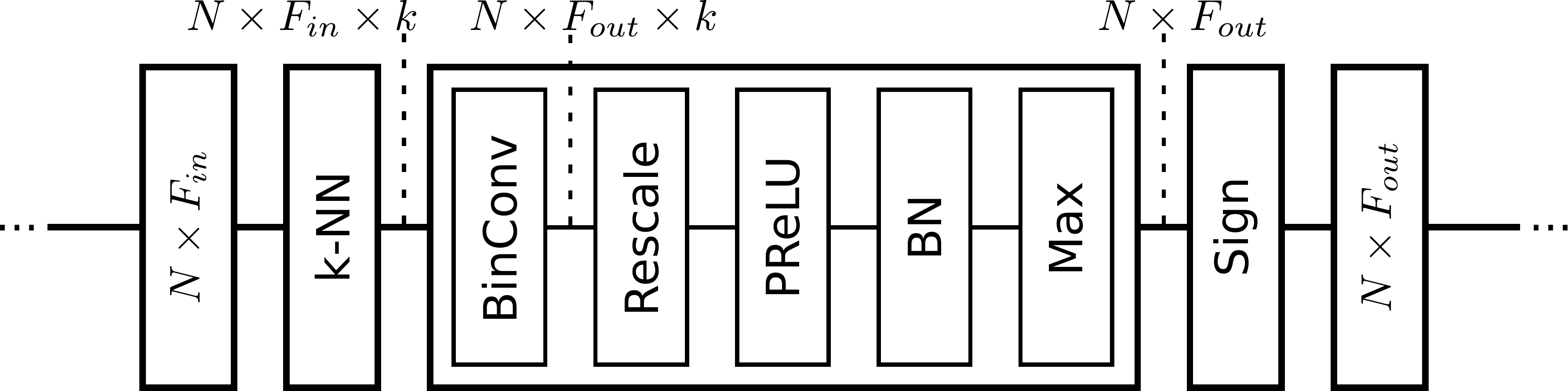}
    \caption{\textbf{Top:} Test accuracy of different binarization schemes at all stages of our cascaded distillation protocol (baseline: $92.89\%$). \textbf{Bottom:} The "BF2" variant of our XorEdgeConv operator.}
    \label{fig:catchy}
\end{figure}

\section{Introduction}

Standard CNNs assume their input to have a regular grid structure, and are therefore suitable for data that can be well-represented in an Euclidean space, such as images, sound, or videos. However, many increasingly relevant types of data do not fit this framework \cite{Bronstein2017}. Graph theory offers a broad mathematical formalism for modeling interactions, and is therefore commonly used in fields such as network sciences \cite{derrible2011applications}, bioinformatics \cite{huang2020skipgnn, 10.3389/fgene.2019.00381}, and recommender systems \cite{mirza2003studying}, as well as for studying discretisations of continuous mathematical structures such as in computer graphics \cite{bronstein_numerical_2008}. This motivates the development of machine learning methods able to deal with graph-supported data. Among them, Graph Neural Networks (GNNs) generalize the operations of CNNs to arbitrary topologies by extending the basic building blocks of CNNs such as convolutions and pooling to graphs. Similarly to CNNs, GNNs learn deep representations of graphs or graph elements, and have emerged as the best performing models for learning on graphs as well as on 3D meshes with the development of advanced and increasingly deep architectures \cite{Li2019,Gong2020}.

As the computational complexity of the networks and the scale of graph datasets increase, so does the need for faster and smaller models. The motivations for resource-efficient deep learning are numerous and also apply to deep learning on graphs and 3D shapes. Computer vision models are routinely deployed on embedded devices, such as mobile phones or satellites \cite{bahl2019low, lemaire2020fpga}, where energy and storage constraints are important. The development of smart devices and IoT may bring about the need for power-efficient graph learning models \cite{khanfor2020graph, zhangequipment, Casas2020SpAGNNSG}. Finally, models that require GPUs for inference can be expensive to serve, whereas CPUs are typically more affordable. This latter point is especially relevant to the applications of GNNs in large-scale data mining on relational datasets, such as those produced by popular social networks, or in bioinformatics \cite{zitnik2018biosnap}.

While recent work has proposed algorithmic changes to make graph neural networks more scalable, such as the use of sampling \cite{Hamilton2017,graphsaint-iclr20} or architectural improvements \cite{Frasca2020,Chiang2019} and simplifications \cite{Wu2019}, our approach is orthogonal to these advances and focuses on compressing existing architectures while preserving model performance. Model compression is a well-researched area for Euclidean neural networks, but has seen very little application in geometric deep learning. In this paper, we study different strategies for binarizing GNNs.

Our contributions are as follows:
\begin{itemize}
\item We present a binarization strategy inspired by the latest developments in binary neural networks for images \cite{DBLP:conf/bmvc/BulatT19,real2binICLR20} and knowledge distillation for graph networks
\item We develop an efficient dynamic graph neural network model that constructs the dynamic graph in Hamming space, thus paving the way for significant speedups at inference time, with negligible loss of accuracy when using real-valued weights
\item We conduct a thorough ablation study of the hyperparameters and techniques used in our approach
\item We demonstrate real-world acceleration of our models on a budget ARM device
\end{itemize}

\mypar{Notations} Matrices and vectors are denoted by upper and lowercase bold letters (\eg, $\mathbf{X}$ and  $\mathbf{x}$), respectively.  
$\mathbf{I}$ denotes the identity matrix of compatible dimensions. The $i^{th}$ column of $\mathbf{X}$ is denoted as $\mathbf{x}_{i}$. The set of real numbers is denoted by $\mathbb{R}$. A {\em graph} $\mathcal{G} = (\mathcal{V}, \mathcal{E})$ consists of 
{\em vertices} $\mathcal{V}=\{1,\hdots, n\}$ and {\em edges} $\mathcal{E} \subseteq \mathcal{V}\times \mathcal{V}$. 
The {\em neighborhood} of vertex $i$, denoted by $\mathcal{N}(i) = \{j : (i,j) \in \mathcal{E} \}$, is the set of vertices adjacent to $i$. %
Other mathematical notations are summarized in Appendix E of the Supplementary Material.

\section{Related Work}

\paragraph*{Knowledge distillation} uses a pretrained \textit{teacher} network to supervise and inform the training of a smaller \textit{student} network
. In logit matching \cite{44873}, a cross-entropy loss is used to regularize the output logits of the student by matching them with a blurred version of the teacher's logits computed using a softmax with an additional temperature hyperparameter. More recent works also focus on matching internal activations of both networks, such as attention volumes in \cite{Zagoruyko2017AT}, or on preserving relational knowledge \cite{Park2019,Lee2019,Tung2019}.

\paragraph*{Quantized and Binary Neural Networks}
Network quantization \cite{gong2014compressing, zhou2016dorefa} refers to the practice of lowering the numerical precision of a model in a bid to reduce its size and speed-up inference. Binary Neural Networks (BNNs) \cite{NIPS2016_d8330f85} push it to the extreme and use a single bit for weights and activations. The seminal work of XNOR-Net \cite{Rastegari2016} showed that re-introducing a small number of floating point operations in BNNs can drastically improve the performance compared to using pure binary operations by reducing the quantization error. In XNOR-Net, a dot product $\star$ between real vectors $\mathbf{a}$ and $\mathbf{b}$ of dimension $n$ is approximated by $\mathbf{a} \star \mathbf{b} \approx (\sign(\mathbf{a}) \circledast \sign(\mathbf{b})) \alpha \beta$, where $\beta = \frac{1}{n}||\mathbf{a}||_1$ and $\alpha = \frac{1}{n}||\mathbf{b}||_1$ are rescaling constants. XNOR-Net++ \cite{DBLP:conf/bmvc/BulatT19} proposed to instead learn a rescaling tensor $\Gamma$, with shared factors to limit the number of trainable parameters and avoid overfitting. Finally, in Real-to-Binary networks \cite{real2binICLR20}, the authors compile state-of-the-art techniques and improve the performance of binary models with knowledge distillation.

\paragraph*{Graph Neural Networks}

Graph Neural Networks were initially proposed in \cite{Gori2005,Scarselli2009} as a form of recursive neural networks. Later formulations relied on Fourier analysis on graphs using the eigendecomposition of the graph Laplacian \cite{Bruna2013b} and approximations of such \cite{defferrard_convolutional_2016}, but suffered from the connectivity-specific nature of the Laplacian. Attention-based models \cite{Monti2017,Fey2017,Verma2018,Velickovic2017} are purely spatial approaches that compute a vertex's features as a dynamic weighting of its neighbours'. Spatial and spectral approaches have been unified \cite{Kipf2017} and shown to derive from the more general neural message passing \cite{Gilmer2017} framework. We refer to recent reviews on GNNs, such as \cite{Wu2020}, for a comprehensive overview, and focus only on the operators we binarize in this paper. 

The message-passing framework offers a general formulation of graph neural networks:
\begin{align}
\begin{split}
    \label{eq:mpnn}
     & \vx_i^{(l)}  = \\
     & \gamma^{(l)}\left( \vx_i^{(l-1)}, \displaystyle \mathop{\square}_{j \in \mathcal{N}(i)} \phi^{(l)}\left(\vx_i^{(l-1)}, \vx_j^{(l-1)}, \vv{e}_{ij}^{(l-1)}\right) \right),
\end{split}
\end{align} where $\square$ denotes a differentiable symmetric (permutation-invariant) function, (\eg $\max$ or $\sum$), $\phi$ a differentiable kernel function, $\gamma$ is an MLP, and $\mathbf{x}_i$ and $\vv{e}_{ij}$ are features associated with vertex $i$ and edge $(i,j)$, respectively.

The EdgeConv operator is a special case introduced as part of the Dynamic Graph CNN (DGCNN) model \cite{Wang2019} and defines an edge message as a function of $\mathbf{x}_j - \mathbf{x}_i$:
\begin{align}
    \mathbf{e}^{(l)}_{ij} & = \text{ReLU} \left( \bm{\theta}^{(l)} (\mathbf{x}^{(l-1)}_j - \mathbf{x}^{(l-1)}_i) + \bm{\phi}^{(l)} \mathbf{x}^{(l-1)}_i \right)\\
              & = \text{ReLU} \left( \bm{\Theta}^{(l)} \tilde{\mathbf{X}}^{(l-1)} \right) \label{eq:edgeconv}
\end{align} where $\tilde{\mathbf{X}}^{(l-1)} = \left[ \mathbf{x}^{(l-1)}_i \concat \mathbf{x}^{(l-1)}_j - \mathbf{x}^{(l-1)}_i \right]$, $\bm{\theta}$ and $\bm{\phi}$ are trainable weights, and $\bm{\Theta}$ their concatenation.

The output of the convolution is the max aggregation ($\square = \max$) of the edge messages:
\begin{equation}
    \mathbf{x}^{(l)}_{i} = \max_{j \in \mathcal{N}(i)} \mathbf{e}^{(l)}_{ij}
\end{equation}
While the EdgeConv operator is applicable to graph inputs, the main use case presented in \cite{Wang2019} is for point clouds, where the neighbours are found by $k$-Nearest Neighbours ($k$-NN) search in feature space before each convolutional layer. DGCNN is the first example of a dynamic graph architecture, with follow-up work in \cite{Kazi2020}.

The GraphSAGE \cite{Hamilton2017} operator introduced inductive learning on large graphs with sampling and can also be phrased as a message passing operator:
\begin{align}
    \mathbf{x}^{(l)}_i = \normalize \left(\text{ReLU} \left( \mathbf{W}^{(l)} \left[ \mathbf{x}^{(l-1)}_i \concat \aggr_{j \in \mathcal{N}(i)} \mathbf{x}^{(l-1)}_j  \right] \right) \right)
    \label{eq:graphsage}
\end{align}
Where $\aggr$ is a symmetric aggregation function such as $\max$, $\mathrm{sum}$ or $\mathrm{mean}$; $\normalize$ denotes the $\ell_2$ normalization, and $\mathbf{W}$ is a tensor of learnable weights.

\paragraph*{Model Compression in Geometric Deep Learning}
In \cite{Wang2020}, the authors propose to binarize the Graph Attention (GAT) operator \cite{Velickovic2017}, and evaluate their method on small-scale datasets such as Cora \cite{motl2015ctu} and Pubmed \cite{Kipf2017}. In \cite{Wang2020_bigcn}, the authors apply the XNOR-Net approach to GCN \cite{Kipf2017} with success, but also on small-scale datasets. Finally, \cite{Qin2020} propose to binarize PointNet with tailored aggregation and scaling functions. At the time of writing, the Local Structure Preserving (LSP) module of \cite{Yang_2020_CVPR} is the only knowledge distillation method specifically designed for GNNs. LSP defines local structure vectors $LS_i$ for each node in the graph:
\begin{equation}
    LS_{ij} = \frac{\exp(\text{SIM}(\mathbf{x}_i, \mathbf{x}_j))}{\sum_{k \in \mathcal{N}(I)} \exp(\text{SIM}(\mathbf{x}_i, \mathbf{x}_k))} \label{eq:lsp}
\end{equation} where $\text{SIM}$ denotes a similarity measure, \eg, $||.||_2^2$ or a kernel function such as a Gaussian RBF kernel. The total local structure preserving loss between a student network $s$ and a teacher $t$ is then defined as:
\begin{equation}
    L_{LSP} = \frac{1}{|\mathcal{V}|} \sum_{i \in \mathcal{V}} \sum_{j \in \mathcal{N}^{u}(i)} LS^{s}_{ij} \log \frac{LS^{s}_{ij}}{LS^{t}_{ij}}.
\end{equation} $\mathcal{N}^{u}(i) = \mathcal{N}^{s}(i) \cup \mathcal{N}^{t}(i)$ to support dynamic graph models.

\section{Method}

Eq. \ref{eq:mpnn} is more general than the vanilla Euclidean convolution, which boils down to a single matrix product to quantize. We must therefore choose which elements of Eq. \ref{eq:mpnn} to binarize and how: the node features $\mathbf{x}_i$, the edge messages $\mathbf{e}_{ij}$, and the functions $\square$, $\gamma$ and $\phi$ may all need to be adapted.

\begin{figure*}[t]
    \centering
    \includegraphics[width=\linewidth]{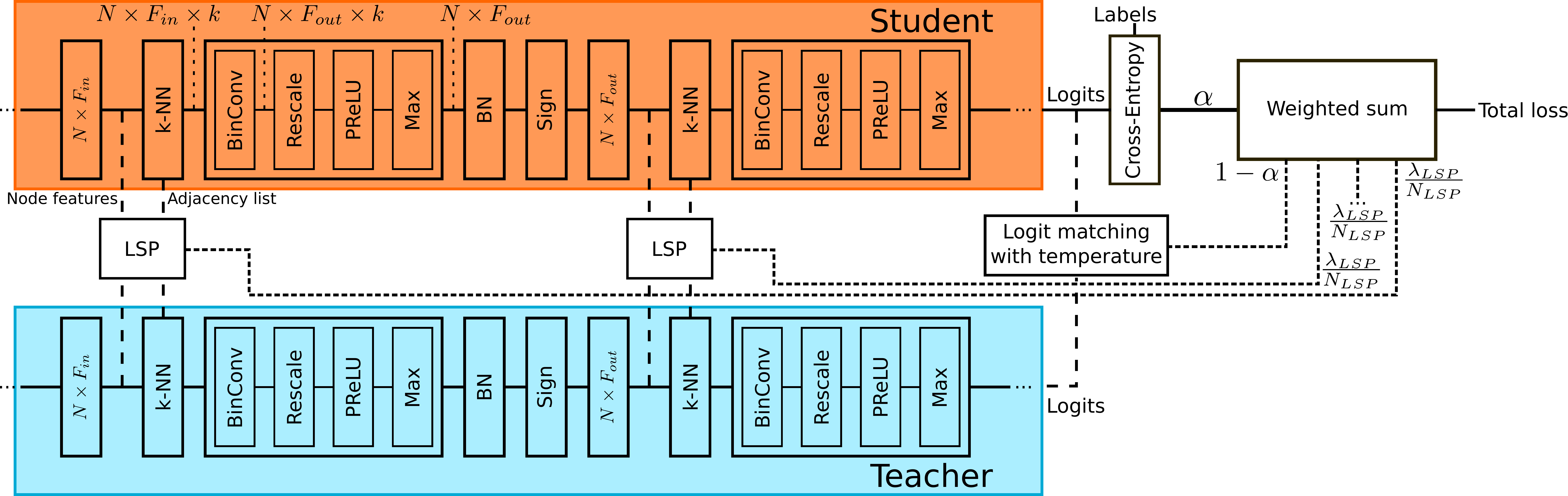}
    \caption{\textbf{Distillation with the "BF1" variant of XorEdgeConv:} the student model is more heavily quantized than the teacher. Knowledge transfer points equipped with LSP modules encourage similar dynamic graph feature distributions after each $k$-NN graph computation (except for the first, performed on the fixed 3D coordinates). Logit matching is used to further inform the training of the student.}
    \label{fig:full_overview}
\end{figure*}

\paragraph*{Quantization} We follow the literature and adopt the $\sign$ operator as the binarization function:
\begin{equation}
    \sign(x) = \begin{cases} 1 &\mbox{if } x \geq 0 \\
                                   -1 & \mbox{if } x < 0 \end{cases}.
                                   \label{eq:sign}
\end{equation}
As the gradient of $\sign$ is zero almost everywhere, we employ the straight-through estimator \cite{bengio2013estimating} to provide a valid gradient. We use this method for both network weights and activations. Furthermore, we mean-center and clip the real latent network weights after their update in the backpropagation step.

\paragraph*{Learnable rescaling}  Assuming a dot product operation (\eg a fully-connected or convolutional layer) $\mathbf{A} \star \mathbf{B} \in \mathbb{R}^{o \times h \times w}$, we approximate it as in \cite{DBLP:conf/bmvc/BulatT19}:
\begin{equation}
    \mathbf{A} \star \mathbf{B} \approx \left( \sign(\mathbf{A}) \circledast \sign(\mathbf{B}) \right) \odot \Gamma,
\end{equation}  with $\Gamma$ a learned rescaling tensor.
We use two constructions of $\Gamma$ depending on the model. Channel-wise:
\begin{equation}
\Gamma = \alpha \in \mathbb{R}^{o \times 1 \times 1}
\label{eq:gamma1}
\end{equation}
and one rank-1 factor per mode:
\begin{equation}
    \Gamma = \alpha \otimes \beta \otimes \gamma, \quad \alpha \in \RR^o, \beta \in \RR^h, \gamma \in \RR^w
    \label{eq:gamma2}
\end{equation} 

\paragraph*{Activation functions} Recent work \cite{Bulat2019} has shown using non-linear activations in XNOR-Net - type blocks can improve the performance of binary neural networks, with PReLU bringing the most improvement. 

\paragraph*{Knowledge Distillation}
Inspired by \cite{real2binICLR20}, we investigate the applicability of knowledge distillation for the binarization of graph neural networks. For classification tasks, we use a logit matching loss \cite{44873} as the base distillation method. We also implemented the LSP module of \cite{Yang_2020_CVPR}.

\paragraph*{Multi-stage training}

We employ a cascaded distillation scheme \cite{real2binICLR20}, an overview of which is shown in Figure \ref{fig:full_overview}.

\noindent \textit{Stage 1}: We first build a real-valued and real-weighted network with the same architecture as the desired binary network by replacing the quantization function with $\tanh$. We distillate the original (base) network into this first student network. We employ weight decay with weight $\lambda = 1e-5$, logit matching, and LSP. We use the same initial learning rate and learning rate schedule as for the base network.

\noindent \textit{Stage 2}: The model of stage 1 becomes the teacher, the student is a binary network with real-valued weights but binary activations. We initialize the student with the weights of the teacher. We employ weight decay with $\lambda = 1e-5$, logit matching, and LSP. We use a smaller learning rate (\eg $25\%$) than for stage 1 and the same learning rate schedule.

\noindent \textit{Stage 3}: The model of stage 2 becomes the teacher, the student is a binary network with binary weights and binary activations. We use logit matching and LSP but no weight decay. The hyperparameters we used are available in Section \ref{sec:implem_details}. We did not observe a significant difference in models initialized randomly or using the weights of the teacher.

\paragraph*{Batch Normalization} We investigate the importance of the order of the dot product and batch normalization operations for discretizing dot product operations within graph convolution operators. However, our base approach is to follow the XNOR-Net block structure \cite{Rastegari2016} with learnable rescaling (\ie XNOR-Net++ block). In particular, all fully-connected layers of MLPs that follow graph feature extraction layers are binarized using the XNOR-Net++ block.

\section{Models}

We choose the Dynamic Graph CNN model, built around the EdgeConv operator of Eq. \ref{eq:edgeconv} as our main case study. DGCNN has several characteristics that make it an interesting candidate for binarization. First, the EdgeConv operator is widely applicable to graphs and point clouds. Second, the operator relies on both node features and edge messages, contrary to other operators previously studied in GNN binarization such as GCN. Third, the time complexity of DGCNN is strongly impacted by the $k$-NN search in feature space.
$k$-NN search can be made extremely efficient in Hamming space, and fast algorithms could theoretically be implemented for the construction of the dynamic graph at inference, provided that the graph features used in the search are binary, which requires a different binarization strategy than merely approximating the dense layer in EdgeConv.

For completeness, we also derive a binary SAGE operator.

\subsection{Direct binarization}
\label{sec:dgcnn_real}

\begin{figure}[t]
    \centering
    \includegraphics[width=\linewidth]{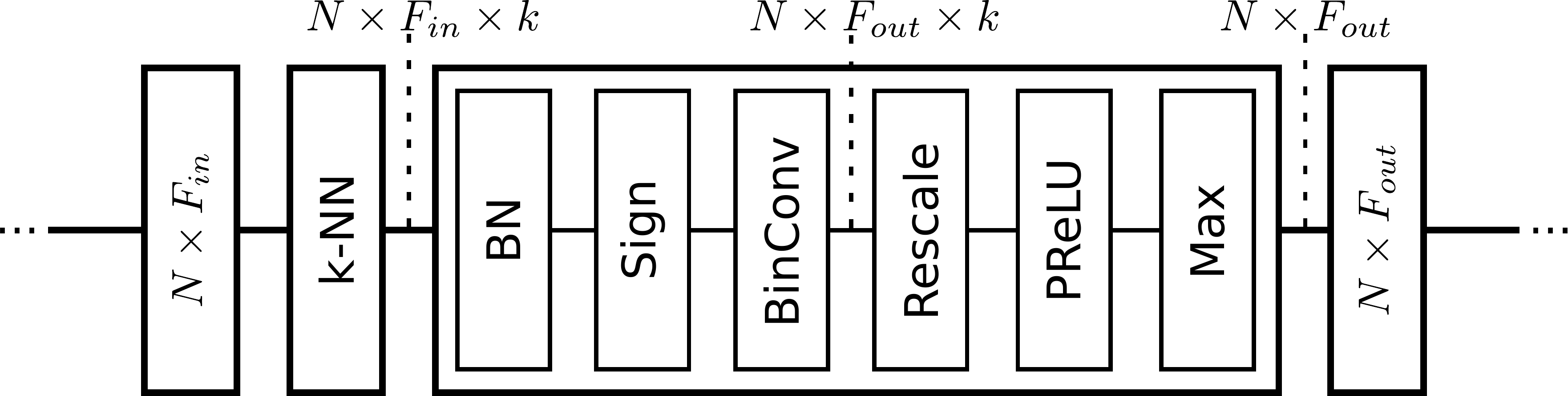}
    \caption{\textbf{The BinEdgeConv operator} ("RF" model in the experiments) can be deployed as a drop-in replacement for EdgeConv and follows the XNOR-Net++ approach to binarization.}
    \label{fig:rf_operator}
\end{figure}

Our first approach binarizes the network weights and the graph features at the input of the graph convolution layers, but keeps the output real-valued. The network, therefore, produces real-valued node features. We replace the EdgeConv operator by a block similar to XNOR-Net++, using learnable rescaling and batch normalization pre-quantization.

We define the BinEdgeConv operator as:
\begin{align}
    \label{eq:binedgeconv}
    \mathbf{e}^{(l)}_{ij} &= \sigma \left(\sign (\bm{\Theta}^{(l)}) \circledast \sign \left( \text{BN}\left( \tilde{\mathbf{X}}^{(l-1)} \right) \right) \odot \Gamma^{(l)} \right)\\
    \mathbf{x}^{(l)}_{i} &= \max_{j \in \mathcal{N}(i)} \mathbf{e}^{(l)}_{ij}
\end{align} with $\sigma$ the PReLU activation, 
and $\Gamma^{(l)}$ a real rescaling tensor. BinEdgeConv is visualized in Figure \ref{fig:rf_operator}.

We use the same structure to approximate the MLP classifier. Similarly, we binarize Eq. \ref{eq:graphsage} to get:
\begin{align}
    \mathbf{h}^{(l)} & = \sign \left( \text{BN} \left( \left[ \mathbf{x}^{(l-1)}_i \concat \aggr_{j \in \mathcal{N}(i)} \mathbf{x}^{(l-1)}_j  \right] \right) \right)\\
    \mathbf{x}^{(l)}_i & = \normalize \left( \sigma \left( ( \sign (\mathbf{W}^{(l)} ) \circledast \mathbf{h}^{(l)} ) \odot \Gamma^{(l)} \right) \right).
    \label{eq:bingraphsage}
\end{align}
with $\sigma$ the PReLu activation and $\Gamma^{(l)}$ following Eq. \ref{eq:gamma1}.

\subsection{Dynamic Graph in Hamming Space}
\label{sec:dgcnn_hamming}

As mentioned, one advantage of binary node features is to enable fast computation of the $k$-Nearest Neighbours graph at inference time by replacing the $\ell_2$ norm with the Hamming distance. We detail our approach to enable quantization-aware training with $k$-NN search on binary vectors.

\paragraph*{Edge feature}
The central learnable operation of EdgeConv is $\bm{\Theta} \left[ \mathbf{x}_i \concat \mathbf{x}_j - \mathbf{x}_i \right]$ as per Eq. \ref{eq:edgeconv}, where the edge feature is $\mathbf{x}_j - \mathbf{x}_i$. Assuming binary node features, the standard subtraction operation becomes meaningless. Formally, for $\mathbf{x}_1, \mathbf{x}_2 \in \RR^n$ with $\RR^n$ the $n$-dimensional Euclidean vector space over the field of real numbers,
\begin{equation}
    \mathbf{x}_1 - \mathbf{x}_2 \coloneqq \mathbf{x}_1 + (-\mathbf{x}_2) \label{eq:addinverse}
\end{equation} 
by definition, with $(-\mathbf{x}_2)$ the additive inverse of $\mathbf{x}_2$. Seeing binary vectors as elements of vector spaces over the finite field $\mathbb{F}_2$, we can adapt Eq. \ref{eq:addinverse} with the operations of Boolean algebra. The addition therefore becomes the Boolean exclusive or (XOR) $\oplus$, and the additive inverse of $(-x)_{\mathbb{F}_2}$ is $x$ itself ($x \oplus x = 0$). With our choice of quantizer (Eq. \ref{eq:sign}), $\mathbf{x}_i, \mathbf{x}_j \in \{-1, 1\}^n$ and we observe that $\mathbf{x}_i \oplus \mathbf{x}_j = - \mathbf{x}_i \, \odot \, \mathbf{x}_j$. We therefore base our binary EdgeConv operator for binary node features, XorEdgeConv, on the following steps:
\begin{align}
    \label{eq:xoredgefeature}
    \mathbf{e}^{(l)}_{ij} & = \sigma \left(\sign (\bm{\Theta}^{(l)}) \circledast \tilde{\mathbf{X}}_b^{(l-1)} \odot \Gamma^{(l)} \right)\\
    \mathbf{x}^{(l)}_{i} & = \sign \left( \max_{j \in \mathcal{N}(i)} \mathbf{e}^{(l)}_{ij} \right)
\end{align} with $\tilde{\mathbf{X}}_b^{(l-1)} = \left[ \mathbf{x}^{(l-1)}_i \concat - \mathbf{x}^{(l-1)}_j \odot \mathbf{x}^{(l-1)}_i \right]$, $\bm{\Theta}^{(l)}$ a set of learnable real parameters and $\Gamma^{(l)}$ a real rescaling tensor. We further investigate the practical importance of the placement of the batch normalization operation, either before or after the aggregation function, by proposing two variants:
\begin{equation}
    \mathbf{x}^{(l)}_{i} = \sign \left( \text{BN} \left( \max_{j \in \mathcal{N}(i)} \mathbf{e}^{(l)}_{ij} \right) \right)
    \label{eq:bf1}
\end{equation} shown as part of Figure \ref{fig:full_overview} and
\begin{equation}
    \mathbf{x}^{(l)}_{i} = \sign \left( \max_{j \in \mathcal{N}(i)} \text{BN} \left( \mathbf{e}^{(l)}_{ij} \right) \right)
    \label{eq:bf2}
\end{equation} 
drawn in Figure \ref{fig:catchy}. Here, the main difference lies in the distribution of the features pre-quantization.

\paragraph*{Nearest Neighbours Search}
 The Hamming distance between two binary vectors $\mathbf{x}, \mathbf{y}$ is $d_H(\mathbf{x}, \mathbf{y}) = ||\mathbf{x} \oplus \mathbf{y}||_H$ where $||.||_H$ is the number of non-zero bits, and can be efficiently implemented as \texttt{popcount(x xor y)}. We note that this relates our approach to previous work on efficient hashing \cite{Norouzi2011,Norouzi2014,8463534} and metric learning \cite{Norouzi2012}, especially given the dynamic nature of the graph. Unfortunately, like the $\sign$ function, the Hamming distance has an ill-defined gradient, which hinders its use as-is for training. We therefore investigate two continuous relaxations. (1) we use the standard $\ell_2$ norm for training, since all norms are equivalent in finite dimensions. (2) we observe that the matrix of pairwise Hamming distances between $d$-dimensional vectors $\mathbf{x}_i$ valued in $\{-1, 1\}$ can be computed in a single matrix-matrix product up to a factor 2 as (see Eq. 5 of \cite{Liu2012}):
\begin{equation}
    \mathbf{D} = -(\mathbf{X} \mathbf{X}^T - d \mathbf{I}_d)
\end{equation} with $\mathbf{X}$ the matrix of the $\mathbf{x}_i$ stacked row-wise, and $\mathbf{I}_d$ the identity matrix.
We investigate both options.

\paragraph*{Local structure} With binary node features, we now have to choose how to define the local structure similarity measure of Eq. \ref{eq:lsp}. One option is to use the standard Gaussian RBF as in the real-valued case. Another option is to define the similarity in Hamming space, like for the $k$-NN search. We therefore investigate the following similarity metric:
\begin{align}
    \label{eq:hamming_sim}
    \text{SIM}(\mathbf{x}_i, \mathbf{x}_j) &= e^{-||\mathbf{x}_i \oplus \mathbf{x}_j||_H}
\end{align}
For vectors $\mathbf{x}, \mathbf{y} \in \{-1, 1\}^n$, we note that $||\mathbf{x} \oplus \mathbf{y}||_H = \frac{1}{2} \sum_{k=1}^n (-x_k y_k + 1)$.

\section{Experimental Evaluation}

We perform a thorough ablation study of our method on Dynamic Graph CNN. The model binarized according to the method of Section \ref{sec:dgcnn_real} and using the BinEdgeConv operator of Eq.~\ref{eq:binedgeconv} is referred to as \textbf{RF} for "Real graph Features". The model binarized according to Section \ref{sec:dgcnn_hamming} and using the XorEdgeConv operator is referred to as \textbf{BF1}, if following Eq. \ref{eq:bf1}, or \textbf{BF2}, if following Eq. \ref{eq:bf2}.
We evaluate DGCNN on the ModelNet40 classification benchmark, as in \cite{Wang2019}. We implement $\Gamma$ as per Eq. \ref{eq:gamma2} for the \textbf{RF} model and Eq. \ref{eq:gamma1} for the \textbf{BF} models. In the next paragraphs, the numbers in parentheses refer to the corresponding lines of Table \ref{tab:big_ablation_table}.

\begin{table*}[t]
\small
    \centering
    \begin{tabular}{c|c|c|c|c|c|c|c|c|c|c}
\# & Model     & Distillation  & Stage & $k$-NN       & LSP       & $\lambda_{LSP}$   & Activation    & Global balance  & Edge balance  & Acc (\%)\\\shline
1  & BF1        & None          & -     & H         & -         & -                 & PReLU         & -               &  -            & 61.26\\
2  & BF1        & Direct        & -     & H         & -         & -                 & PReLU         & -               &  -            & 62.84\\
3  & BF1        & Cascade   & 3     & H         & -         & -                 & PReLU         & -               &  -            & \textbf{81.04}\\
4  & BF1        & Cascade   & 3     & H         & -         & -                 & None          & -               &  -            & 74.80\\
\hline
5  & BF1        & Cascade   & 2     & H         & $\ell_2$  & 100               & PReLU         & -               &  -            & \textit{90.32}\\
6  & BF1        & Cascade   & 3     & H         & $\ell_2$  & 100               & PReLU         & -               &  -            & \textbf{81.32}\\
\hline
7  & BF1        & Cascade   & 3     & H         & $\ell_2$  & 100               & ReLU          & -               &  -            & 76.46\\
\hline
8  & BF1        & None          & -     & $\ell_2$  & -         & -                 & PReLU         & -               &  -            & 62.97\\
9  & BF1        & Direct        & -     & $\ell_2$  & -         & -                 & PReLU         & -               &  -            & 61.95\\
10  & BF1        & Cascade   & 3     & $\ell_2$  & -         & -                 & PReLU         & -               &  -            & \textbf{81.00}\\
\hline
11 & BF1        & Cascade   & 2     & $\ell_2$  & $\ell_2$  & 100               & PReLU         & -               &  -            & \textit{91.00}\\
12 & BF1        & Cascade   & 3     & $\ell_2$  & $\ell_2$  & 100               & PReLU         & -               &  -            & \textbf{81.00}\\
\shline
13 & BF2        & None          & -     & H         & -         & -                 & PReLU         &   -              &  -            & 59.72\\
14 & BF2        & Direct        & -     & H         & -         & -                 & PReLU         &   -              &  -            & 59.00\\
15 & BF2        & Cascade   & 3     & H         & -         & -                 & PReLU         &       -          &  -            & \textbf{79.58}\\
\hline
16 & BF2        & Cascade   & 3     & H         & -         & -                 & PReLU         & -               &  Mean         & 51.05\\
17 & BF2        & Cascade   & 3     & H         & -         & -                 & PReLU         & -               &  Median       & \textbf{75.93}\\
18 & BF2        & Cascade   & 3     & H         & -         & -                 & None          & -               &  Median       & 71.96\\
\hline
19 & BF2        & Cascade   & 2     & H         & $\ell_2$  & 100               & PReLU         & -               &  -            & \textit{91.57}\\
20 & BF2        & Cascade   & 3     & H         & $\ell_2$  & 100               & PReLU         & -               &  -            & \textbf{81.08}\\
\hline
21 & BF2        & Cascade   & 3     & H         & $\ell_2$  & 100               & None          & -               &  -            & 76.09\\
22 & BF2        & Cascade   & 3     & H         & $\ell_2$  & 100               & ReLU          & -               &  -            & \textbf{76.22}\\
\hline
23 & BF2        & Cascade   & 3     & H         & $\ell_2$  & 100               & PReLU         & Mean            &  -            & \textbf{67.87}\\
24 & BF2        & Cascade   & 3     & H         & $\ell_2$  & 100               & PReLU         & Median          &  -            & 60.82\\
\hline
25 & BF2        & None          & -     & $\ell_2$  & -         & -                 & PReLU         & -               &  -            & 57.90\\
26 & BF2        & Direct        & -     & $\ell_2$  & -         & -                 & PReLU         & -               &  -            & 59.12\\
27 & BF2        & Cascade   & 3     & $\ell_2$  & -         & -                 & PReLU         & -               &  -            & \textbf{80.11}\\
\hline
28 & BF2        & Cascade   & 2     & $\ell_2$  & $\ell_2$  & 100               & PReLU         & -               &  -            & \textit{91.53}\\
29 & BF2        & Cascade   & 3     & $\ell_2$  & $\ell_2$  & 100               & PReLU         & -               &  -            & \textbf{81.52}\\
\shline
30 & RF         & None          & -     & $\ell_2$  & -         & -                 & PReLU         & -               &  -            & 79.30\\
31 & RF         & Direct        & -     & $\ell_2$  & -         & -                 & PReLU         & -               &  -            & 72.69\\
32 & RF         & Cascade   & 3     & $\ell_2$  & -         & -                 & PReLU         & -               &  -            & \textbf{91.05}\\
\hline
33 & RF         & Cascade   & 3     & $\ell_2$  & $\ell_2$  & 100               & PReLU         & -               &  -            & \textbf{90.52}\\
34 & RF         & Cascade   & 3     & $\ell_2$  & $\ell_2$  & 100               & None          & -               &  -            & 89.71\\
35 & RF         & Cascade   & 3     & $\ell_2$  & $\ell_2$  & 100               & ReLU          & -               &  -            & 89.59\\
\shline
36 & Baseline & - & - & - & - & ReLU & - & - & - & 92.89\\
    \end{tabular}
    \caption{Different variants and ablations of our binarized DGCNN models on the ModelNet40 dataset.}
    \label{tab:big_ablation_table}
\end{table*}

\paragraph*{Balance functions (16-18, 23-24)} Recent work \cite{Qin2020} has uncovered possible limitations in binary graph and point cloud learning models when quantizing the output of the max-pooling aggregation of batch-normalized high-dimensional features. Similarly, the authors of \cite{Wang2020} claim that a balance function is necessary to avoid large values in the outputs of the dot product operations when most pre-quantization inputs are positive. 

We evaluate two strategies for re-centering the input of $\sign$
, namely mean-centering, and median-centering (thus ensuring a perfectly balanced distribution of positive and negative values pre-quantization). We evaluate these techniques for the max aggregation of edge messages ("edge balance", \eg between the framed block and the $\sign$ operation in Figure \ref{fig:catchy}) and for the max and average pooling operations before the MLP classifier ("global balance").

We can see in Table \ref{tab:big_ablation_table} that in all cases, the addition of balance functions actually lowered the performance of the models. This suggests that using batch normalization prior to quantization, as is common in the binary CNN literature, is sufficient at the message aggregation level and for producing graph embedding vectors.

\paragraph*{Non-linear activation (3-4,6-7,17-18,20-22,33-35)}

Since the $\sign$ operation can be seen as acting as an activation applied on the output and to the weights of the XorEdgeConv operator, we compare the models with binary node features with PReLU, ReLU, or no additional activation in Table \ref{tab:big_ablation_table}. We can see the PReLU non-linearity offers significant improvements over the models trained with ReLU or without non-linearity in the edge messages, at the cost of a single additional \texttt{fp32} parameter - the largest improvement being observed for the models that apply either median-centering or batch normalization before the quantization operation.

\paragraph*{Binary node features and $k$-NN}

We now study the final performance of our models depending on whether we use BinEdgeConv (real node features) or XorEdgeConv. Looking at the final models (stage 3) in Table \ref{tab:big_ablation_table}, the model with real-valued node features that performs $k$-NN search with the $\ell_2$ norm (32) performs comparably with the full floating-point model (36). On the other hand, we saw a greater reduction in accuracy with the binary node features for the full binary models (6,12,20,29), and comparable accuracy whether we use the $\ell_2$ norm (12,29) or the relaxed Hamming distance (6,20). However, as reported in Table \ref{tab:big_ablation_table}, using real weights (stage 2) with binary node features and $k$-NN search performed in Hamming space (5,11,19,28) matched the performance of the original floating point model (36). 

We found stage 3 to be crucial to the final model's performance, and sensitive to the choice of learning rate and learning rate schedule. This suggests that, although more research and parameter tuning is necessary to maximize the performance of the full binary networks in Hamming space, dynamic graph networks that learn binary codes and construct the dynamic graph in Hamming space can be trained with minimal reduction in performance.

\paragraph{Impact of LSP}

The node features of the teacher and of the students are always real-valued at stage 1. Stage 2 was carried out using either the Gaussian RBF similarity or Eq. \ref{eq:hamming_sim} for the student (which may have binary node features) and the Gaussian RBF for the teacher. Stage 3 uses either similarity measure for both the teacher and student. We also report the results of distilling the baseline DGCNN (full floating-point) model into a BF1 or BF2 full-binary model using the similarity in Hamming space for the student.

We saw inconsistent improvements
when using LSP with the Gaussian RBF ($\ell_2$), as seen in Table \ref{tab:big_ablation_table} (3,6,10,12,15,20,27,29,32,33). This suggest the usefulness of the additional structure preserving knowledge is situational, as it can both increase (3,4,15,20,27,29) or decrease model performance (32,33). Contrary to the models trained using $k$-NN search performed in Hamming space, the models trained with distillation using Eq. \ref{eq:hamming_sim} did not match the performance of their Gaussian $\ell_2$ counterparts, as shown in Table \ref{tab:abl_distillation}, which we conjecture to be due to poor gradients.

\begin{table}[h]
\small
    \centering
    \begin{tabular}{c|c|c|c|c|c}
Model & Stage   & KNN & LSP & $\lambda_{LSP}$ & Acc. (\%)\\
\shline
BF1   & 2       & H         & H & 100 & 38.21\\
BF1   & 2       & $\ell_2$  & H & 100 & 38.94\\
BF2   & 2       & H         & H & 100 & 63.25\\
BF2   & 2       & $\ell_2$  & H & 100 & 64.71\\
\hline
BF1   & 3       & H         & H & 100 & 16.29\\
BF1   & 3       & $\ell_2$  & H & 100 & 20.34\\
BF2   & 3       & $\ell_2$  & H & 100 & 9.40\\
BF2   & 3       & H         & H & 100 & 11.47\\
\hline
BF1   & Direct       & $\ell_2$  & H & 100 & 23.34\\
BF2   & Direct       & H         & H & 100 & 30.23\\
BF2   & Direct       & $\ell_2$  & H & 100 & 32.17\\
BF1   & Direct       & H         & H & 100 & 36.47\\
    \end{tabular}
    \caption{Performance of models trained with LSP using the Hamming-based similarity of Eq. \ref{eq:hamming_sim} (H) at different stages and for direct distillation. Compared to the models trained using the Gaussian RBF ($\ell_2$) similarity, low performance was observed.}
    \label{tab:abl_distillation}
\end{table}

\paragraph*{Cascaded distillation (1-3,13-15,25-27,30-32)}

Table \ref{tab:big_ablation_table} compares distilling the baseline network directly into a binary network, training from scratch, and the three-stage distillation. We observed consistently higher performance with the progressive distillation, confirming its effectiveness.

\paragraph{Large-scale inductive learning with GraphSAGE}

We benchmark our binarized GraphSAGE on the 
OGB-Products and OGB-Proteins node property prediction datasets \cite{hu2020open}, which are recent and
challenging (2,449,029 nodes, 61,859,140 edges for OGB-Product) benchmarks with standardized evaluation procedures, compared to the more commonly used ones, such as 
Cora \cite{motl2015ctu} (2708 nodes, 5429 edges) used in \cite{Wang2020} or Reddit \cite{Hamilton2017} (232,965 nodes, 114,615,892 edges) used in \cite{Wang2020_bigcn}. Notably, the Proteins dataset is challenging due to the high average node degree and small graph diameter, which may exacerbate limitations of GNNs \cite{Alon2020}.

\begin{table}[h]
    \centering
    \resizebox{\columnwidth}{!}{
    \begin{tabular}{c|c|c|c|c}
            & \multicolumn{2}{c}{OGBN-Products} & \multicolumn{2}{c}{OGBN-Proteins}\\
        Model & Mean acc. & Std. acc. & Mean acc. & Std. acc.\\
        \shline
        SAGE fp32 &  0.7862 & 0.0043 & 0.7734 & 0.0041\\
        SAGE bin f.s. & 0.7300 & 0.0156 & 0.7497 & 0.0047\\
        SAGE bin l.m. & 0.7260 & 0.0153 & - & -\\
        GCN fp32 & 0.7564 & 0.0021 & 0.7254 & 0.0044\\
    \end{tabular}
    }
    \caption{Final test accuracy on the OGB-Products and OGB-Proteins node property prediction benchmarks, averaged over 10 runs. "f.s.": from scratch. "l.m.": logit matching.}
    \label{tab:my_label}
\end{table}

We implemented BinSAGE according to Eq.~\ref{eq:bingraphsage} - details of the architecture can be found in Section \ref{sec:implem_details}. For OGB-Products, we use logit matching only for distillation and no PReLU activation. For OGB-Proteins, we use PReLU activations and no distillation, as the task is multi-label classification, and the very large number of edges made using LSP impractical. We use channel-wise rescaling only for both to maximize scalability. On OGB-Products, we did not observe a statistically significant different between training the model from scratch and three-stage distillation with logit matching: in both cases, the full binary model came within 5-6\% of the full-precision model. On OGB-Proteins, the simple binary network trained from scratch is within 3\% of the accuracy of the full-precision network and outperforms the full-precision GCN. This suggests other strategies to improve model scalability, in this case sampling, can be successfully combined with our binarisation method.

\subsection{Speed on embedded hardware }

In order to measure the speed improvements yielded by
our binary conversion scheme, we benchmark it
on a Raspberry Pi 4B board with 4GB of RAM and a Broadcom BCM2711 Quad core Cortex-A72 (ARM v8) 64-bit SoC clocked at 1.5GHz, running Manjaro 64-bit. 
The Pi is a popular, cheap, and readily available ARM-based platform, and is thus a good fit for our experiments.

We benchmark four DGCNN models, in order to measure the speedup
for each subsequent optimization. The specifics of each model are
given in Table~\ref{tab:bench_models}. The input size is set to 1024
points with a batch size of 8, 40 output classes, and 20 nearest neighbors. We convert our models to Tensorflow Lite
using LARQ~\cite{Geiger2020},
an open-source library for binarized neural networks,
and benchmark them using the LARQ Compute Engine (LCE) tool.
Once converted, the smallest model's file size is only 341KB down from
7.2MB, for a 20x reduction.

\begin{table}[h!]
\scriptsize
    \centering
    \begin{tabular}{c|c|c|c}
    
        Model & Binary Weights & Binary Features & Hamming Distance \\
        \hline
        DGCNN &                &                 &     \\
        BDGCNN RF &  \checkmark  &                 &    \\
        BDGCNN BF &  \checkmark  &    \checkmark             &      \\
        BDGCNN BF H &  \checkmark  &  \checkmark               &      \checkmark   \\
    \end{tabular}
    \caption{\footnotesize Features of benchmarked models. Hamm Dist = Pairwise Hamming distance instead of $\ell_2$, implemented in ARM NEON operations on bit-packed features (simulated). }
    \label{tab:bench_models}
\end{table}

\begin{figure}[h!]
    \centering
    \includegraphics[width=0.9\columnwidth]{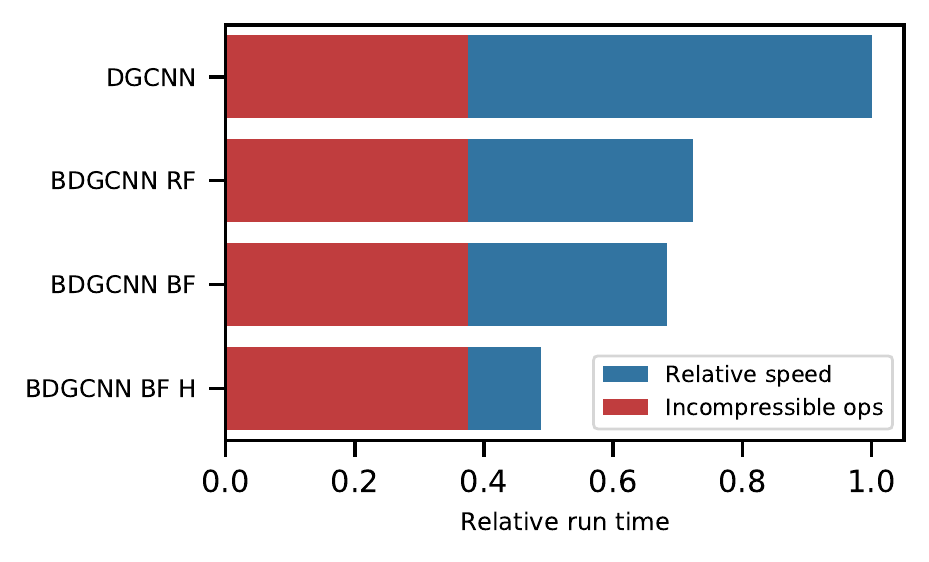}
    \caption{\footnotesize Relative run time on a Raspberry Pi 4B 
    compared to the base DGCNN model. 
    A 2x speedup is achieved by our final optimized model. 
    Run times computed with the LCE benchmark tool over 50 runs.}
    \label{fig:bench_pi}
\end{figure}

Figure~\ref{fig:bench_pi} shows the benchmark results.
Our optimized binary model halves the run-time, thus achieving a substantial speedup.
Peak memory usage is also significantly reduced, from 575MB to 346MB. It is to be noted that DGCNN is a complex model with costly operations, such as concatenation and top-$k$,  that are not made faster by binarization (denoted as "incompressible ops" in Figure \ref{fig:bench_pi}). We provide a profiling of the models in Appendix B of the Supplementary Material.

Unfortunately, we did not have an optimized version of the Hamming distance 
in LCE at the time of writing. Thus, the final 
result is simulated by profiling the run-time of an un-optimized implementation, and estimating the savings we would get with ARM NEON instructions.
It is theoretically possible to treat 64 features at a time, and achieve a 64x speedup (or higher by grouping loads and writes with \texttt{vld4}). We use 32x as a \textit{conservative estimate} since we couldn't account for LCE's bit-packed conversion.

\subsection{Implementation details}
\label{sec:implem_details}

For DGCNN, we follow the architecture of \cite{Wang2018_dgcnn}. For GraphSAGE, we use the baseline architecture of the OGB benchmarks \cite{hu2020open}; that is, three layers with 256 hidden features and mean aggregation. We use three knowledge transfer points for LSP on DGCNN, one after each EdgeConv layer except for the first layer (the $k$-NN and graph features are computed on the 3D coordinates of the point clouds and do not change). All binary models use binary inputs for the convolution and dense layers. For DGCNN, the final layer of the MLP classifier is kept real-weighted, as is customary in the binary neural network literature due to the small number of parameters, but the input features are binarized. For GraphSAGE, all three layers have binary weights.

Our models are implemented in Pytorch \cite{NEURIPS2019_9015}. We use the implementation of DGCNN by the authors as a starting point, and Pytorch Geometric \cite{Fey/Lenssen/2019} for GraphSAGE and OGB \cite{hu2020open}. We use the Adam optimizer \cite{Kingma2014}. We train the DGCNN models for 250 epochs for stage 1 and 350 epochs for stages 2 and 3, on 4 Nvidia 2080 Ti GPUs. The intial learning rate of stage 1 is set to $1e-3$ and for stage 2 to $2.5e-4$, with learning rate decay of $0.5$ at $50\%$ and $75\%$ of the total number of epochs. For stage 3, we set the learning rate to $1e-3$ and decay by a factor of $0.5$ every 50 epochs. We trained GraphSAGE according to the OGB benchmark methodology, using the provided training, validation, and test sets. We trained all models for 20 epochs and averaged the performance over 10 runs. For GraphSAGE, we used $\ell_2$ regularization on the learnable scaling factors only, with a weight $\lambda=1e-4$. For logit matching, we set $T = 3$ and $\alpha = 1e-1$. For LSP, we set $\lambda_{LSP} = 1e2$.

\section{Conclusion}

In this work, we introduce a binarization scheme for GNNs based on the XNOR-Net++ methodology and knowledge distillation. We study the impact of various strategies and design decisions on the final performance of binarized graph neural networks, and show that our approach allows us to closely match or equal the performance of floating-point models on difficult benchmarks, with significant reductions in memory consumption and inference time. We further demonstrate that dynamic graph neural networks can be trained to high accuracy with binary node features, enabling fast construction of the dynamic graph at inference time through efficient Hamming-based algorithms, and further relating dynamic graph models to metric learning and compact hashing. Our DGCNN in Hamming space nearly equals the performance of the full floating point model when trained with floating point weights, and offers competitive accuracy with large speed and memory savings when trained with binary weights. We believe higher performance can already be obtained with this model by adjusting the learning rate schedule in the final distillation stage. Future work will investigate further improving the accuracy of the models, theoretical properties on binary graph convolutions, and inference with fast $k$-NN search in Hamming space.

\paragraph*{Acknowledgements} 
M. B. is supported by a Department of Computing scholarship from Imperial College London, and a Qualcomm Innovation Fellowship. G.B. is funded by the CIAR project at IRT Saint-Exupéry. S.Z. was partially funded by the EPSRC Fellowship DEFORM: Large Scale Shape Analysis of Deformable Models of Humans (EP/S010203/1) and an Amazon AWS Machine Learning Research Award. The authors are grateful to the OPAL infrastructure from Université Côte d'Azur for providing resources and support.
This work was granted access to the HPC resources of IDRIS under the allocation 2020-AD011011311R1 made by GENCI.

\appendix

\section{DGCNN and ModelNet40}

In this appendix, we provide details of the DGCNN model and of the ModelNet40 dataset ommitted from the main text for brevity.

\paragraph{ModelNet40 classification} 
The ModelNet40 dataset \cite{Wu2015} contains 12311 shapes representing 3D CAD models of man-made objects pertaining to 40 categories. We follow the experimental setting of \cite{Wang2018_dgcnn} and \cite{Qi2017a}. We keep 9843 shapes for training and 2468 for testing. We uniformly sample 1024 points on mesh faces weighted by surface area and normalize the resulting point clouds in the unit sphere. The original meshes are discarded. Only the 3D cartesian coordinates $(x,y,z)$ of the points are used as input. We use the same data augmentation techniques (random scaling and perturbations) as \cite{Wang2018_dgcnn} and base our implementation on the author's public code\footnote{\scriptsize\texttt{https://github.com/WangYueFt/dgcnn/tree/master/pytorch}}. We report the overall accuracy as the model score.

\paragraph{Model architecture} All DGCNN models use 4 EdgeConv (or BinEdgeConv or XorEdgeConv) layers with 64, 64, 128, and 256 output channels and no spatial transformer networks. According to the architecture of \cite{Wang2018_dgcnn}, the output of the four graph convolution layers are concatenated and transformed to node embeddings of dimension 1024. We use both global average pooling and global max pooling to obtain graph embeddings from all node embeddings; the resulting features are concatenated and fed to a three layer MLP classifier with output dimensions 512, 256, and 40 (the number of classes in the dataset). We use dropout with probability $p = 0.5$.

\section{Low-level implementation}

This appendix provides further details on the low-level implementation and memory cost of our models.

\subsection{Parameter counts}

We report the counts of binary and floating-point parameters for the baseline DGCNN and our binary models (stage 3) in Table \ref{tab:param_count}.

\begin{table}[h]
    \centering
    \begin{tabular}{c|c|c|c}
         Model & FP32 Param.     & Bin. param.     & Total param.\\
         \shline
        Baseline   &    1,812,648    &          0        &  1,812,648\\
        BF1    &      11,064      &       1,804,672     &  1,815,736\\
        BF2    &      11,064      &       1,804,672     &  1,815,736\\
        RF     &      15,243      &       1,804,672     &  1,819,915\\
    \end{tabular}
    \caption{Number of parameters given by torchsummaryX. Separated into FP and binary operations. 99.39\% of the parameters are binary for \textbf{BF1} and \textbf{BF2}, 99.16\% of the parameters are binary for \textbf{RF}.}
    \label{tab:param_count}
\end{table}

As can be seen in Table \ref{tab:param_count}, our binarization procedure introduces a few extra parameters, but over 99\% of the network parameters are binary.

\subsection{Profiling and optimization of DGCNN}

In order to obtain the data from Section 5.1 of the main paper, we
convert our models with the LARQ converter and benchmark them using
the LCE benchmark utility.

The pairwise Hamming distance is naively implemented as a matrix 
multiplication operation (Eq. 21 of the main text), and we obtain the profiler data
in Table \ref{tab:profiler}, where we have highlighted the nodes
used by that operation. However, not all nodes of these types belong
to the three pairwise distances calculations. We thus provide in Table \ref{tab:profiler2}
the complete profiler output for only one distance calculation in binary 
space, of which there are three in the DGCNN models.

\begin{table}[h]
\footnotesize
  \centering
  \begin{tabular}{c|c|c|c}
Node Type & Avg. ms & Avg \% & Times called \\ \hline
TOPK\_V2 & 488.007 & 22.18\% & 4 \\ 
CONCATENATION & 384.707 & 17.485\% & 6 \\ 
\color{red} FULLY\_CONNECTED & 171.175 & 7.77994\% & 32 \\ 
PRELU & 143.086 & 6.50329\% & 7 \\ 
TILE & 136.443 & 6.20137\% & 4 \\ 
LceBconv2d & 127.371 & 5.78904\% & 6 \\ 
MAX\_POOL\_2D & 122.743 & 5.5787\% & 5 \\ 
MUL & 105.993 & 4.81741\% & 11 \\ 
\color{red} SUB & 92.382 & 4.19878\% & 4 \\ 
LceQuantize & 91.168 & 4.14361\% & 10 \\ 
NEG & 78.453 & 3.56571\% & 4 \\ 
\color{red} PACK & 56.301 & 2.55889\% & 4 \\ 
GATHER & 55.989 & 2.54471\% & 4 \\ 
CONV\_2D & 39.096 & 1.77692\% & 2 \\ 
\color{red} RESHAPE & 35.091 & 1.59489\% & 82 \\ 
ADD & 28.557 & 1.29792\% & 6 \\ 
\color{red} TRANSPOSE & 23.829 & 1.08303\% & 36 \\ 
AVERAGE\_POOL\_2D & 8.071 & 0.366829\% & 1 \\ 
\color{red} SLICE & 5.278 & 0.239886\% & 64 \\ 
LceDequantize & 5.174 & 0.235159\% & 4 \\ 
SUM & 1.132 & 0.0514497\% & 1 \\ 
SQUARE & 0.153 & 0.00695389\% & 1 \\ 
SOFTMAX & 0.01 & 0.000454502\% & 1 \\ 
  \end{tabular}
  \caption{LCE Profiler data for "BDGCNN BF H", summary by node types. 
  In red: nodes that appear in Matmul op which can be
  rewritten as NEON operations for Hamming distance.}
  \label{tab:profiler}
\end{table}

\begin{table*}[t]
\footnotesize
  \centering
  \begin{tabular}{c|c|c|c}
Node type & Avg. time & Avg? \% & Operation name \\
TRANSPOSE & 5.91618 & 0.268874\% & [bin\_dgcnn\_b\_f1\_h/MatMul\_315]:208 \\ 
SLICE & 0.45752 & 0.020793\% & [bin\_dgcnn\_b\_f1\_h/MatMul\_316]:209 \\ 
RESHAPE & 0.43792 & 0.0199023\% & [bin\_dgcnn\_b\_f1\_h/MatMul\_317]:210 \\ 
SLICE & 0.14872 & 0.00675892\% & [bin\_dgcnn\_b\_f1\_h/MatMul\_318]:211 \\ 
RESHAPE & 0.22018 & 0.0100066\% & [bin\_dgcnn\_b\_f1\_h/MatMul\_319]:212 \\ 
SLICE & 0.16006 & 0.0072743\% & [bin\_dgcnn\_b\_f1\_h/MatMul\_320]:213 \\ 
RESHAPE & 0.22242 & 0.0101084\% & [bin\_dgcnn\_b\_f1\_h/MatMul\_321]:214 \\ 
SLICE & 0.16268 & 0.00739337\% & [bin\_dgcnn\_b\_f1\_h/MatMul\_322]:215 \\ 
RESHAPE & 0.21126 & 0.0096012\% & [bin\_dgcnn\_b\_f1\_h/MatMul\_323]:216 \\ 
SLICE & 0.15406 & 0.00700161\% & [bin\_dgcnn\_b\_f1\_h/MatMul\_324]:217 \\ 
RESHAPE & 0.20686 & 0.00940123\% & [bin\_dgcnn\_b\_f1\_h/MatMul\_325]:218 \\ 
SLICE & 0.1494 & 0.00678983\% & [bin\_dgcnn\_b\_f1\_h/MatMul\_326]:219 \\ 
RESHAPE & 0.21348 & 0.00970209\% & [bin\_dgcnn\_b\_f1\_h/MatMul\_327]:220 \\ 
SLICE & 0.15514 & 0.00705069\% & [bin\_dgcnn\_b\_f1\_h/MatMul\_328]:221 \\ 
RESHAPE & 0.2117 & 0.00962119\% & [bin\_dgcnn\_b\_f1\_h/MatMul\_329]:222 \\ 
SLICE & 0.15154 & 0.00688708\% & [bin\_dgcnn\_b\_f1\_h/MatMul\_330]:223 \\ 
RESHAPE & 0.17318 & 0.00787056\% & [bin\_dgcnn\_b\_f1\_h/MatMul\_331]:224 \\ 
SLICE & 0.15244 & 0.00692799\% & [bin\_dgcnn\_b\_f1\_h/MatMul\_332]:225 \\ 
RESHAPE & 0.16864 & 0.00766423\% & [bin\_dgcnn\_b\_f1\_h/MatMul\_333]:226 \\ 
SLICE & 0.15614 & 0.00709614\% & [bin\_dgcnn\_b\_f1\_h/MatMul\_334]:227 \\ 
RESHAPE & 0.1736 & 0.00788965\% & [bin\_dgcnn\_b\_f1\_h/MatMul\_335]:228 \\ 
SLICE & 0.15118 & 0.00687072\% & [bin\_dgcnn\_b\_f1\_h/MatMul\_336]:229 \\ 
RESHAPE & 0.17086 & 0.00776513\% & [bin\_dgcnn\_b\_f1\_h/MatMul\_337]:230 \\ 
SLICE & 0.149 & 0.00677165\% & [bin\_dgcnn\_b\_f1\_h/MatMul\_338]:231 \\ 
RESHAPE & 0.16924 & 0.0076915\% & [bin\_dgcnn\_b\_f1\_h/MatMul\_339]:232 \\ 
SLICE & 0.1505 & 0.00683982\% & [bin\_dgcnn\_b\_f1\_h/MatMul\_340]:233 \\ 
RESHAPE & 0.16894 & 0.00767787\% & [bin\_dgcnn\_b\_f1\_h/MatMul\_341]:234 \\ 
SLICE & 0.14994 & 0.00681437\% & [bin\_dgcnn\_b\_f1\_h/MatMul\_342]:235 \\ 
RESHAPE & 0.17058 & 0.0077524\% & [bin\_dgcnn\_b\_f1\_h/MatMul\_343]:236 \\ 
SLICE & 0.15018 & 0.00682528\% & [bin\_dgcnn\_b\_f1\_h/MatMul\_344]:237 \\ 
RESHAPE & 0.16996 & 0.00772422\% & [bin\_dgcnn\_b\_f1\_h/MatMul\_345]:238 \\ 
SLICE & 0.1496 & 0.00679892\% & [bin\_dgcnn\_b\_f1\_h/MatMul\_346]:239 \\ 
RESHAPE & 0.17112 & 0.00777694\% & [bin\_dgcnn\_b\_f1\_h/MatMul\_347]:240 \\ 
TRANSPOSE & 0.41792 & 0.0189933\% & [bin\_dgcnn\_b\_f1\_h/MatMul\_348]:241 \\ 
FULLY\_CONNECTED & 8.78396 & 0.399207\% & [bin\_dgcnn\_b\_f1\_h/MatMul\_349]:242 \\ 
TRANSPOSE & 0.72016 & 0.0327293\% & [bin\_dgcnn\_b\_f1\_h/MatMul\_350]:243 \\ 
FULLY\_CONNECTED & 8.64452 & 0.39287\% & [bin\_dgcnn\_b\_f1\_h/MatMul\_351]:244 \\ 
TRANSPOSE & 0.71804 & 0.032633\% & [bin\_dgcnn\_b\_f1\_h/MatMul\_352]:245 \\ 
FULLY\_CONNECTED & 8.63224 & 0.392312\% & [bin\_dgcnn\_b\_f1\_h/MatMul\_353]:246 \\ 
TRANSPOSE & 0.72162 & 0.0327957\% & [bin\_dgcnn\_b\_f1\_h/MatMul\_354]:247 \\ 
FULLY\_CONNECTED & 8.62624 & 0.392039\% & [bin\_dgcnn\_b\_f1\_h/MatMul\_355]:248 \\ 
TRANSPOSE & 0.68654 & 0.0312014\% & [bin\_dgcnn\_b\_f1\_h/MatMul\_356]:249 \\ 
FULLY\_CONNECTED & 8.6722 & 0.394128\% & [bin\_dgcnn\_b\_f1\_h/MatMul\_357]:250 \\ 
TRANSPOSE & 0.69886 & 0.0317613\% & [bin\_dgcnn\_b\_f1\_h/MatMul\_358]:251 \\ 
FULLY\_CONNECTED & 8.6892 & 0.394901\% & [bin\_dgcnn\_b\_f1\_h/MatMul\_359]:252 \\ 
TRANSPOSE & 0.71076 & 0.0323021\% & [bin\_dgcnn\_b\_f1\_h/MatMul\_360]:253 \\ 
FULLY\_CONNECTED & 8.70248 & 0.395504\% & [bin\_dgcnn\_b\_f1\_h/MatMul\_361]:254 \\ 
TRANSPOSE & 0.71256 & 0.0323839\% & [bin\_dgcnn\_b\_f1\_h/MatMul\_362]:255 \\ 
FULLY\_CONNECTED & 8.76456 & 0.398326\% & [bin\_dgcnn\_b\_f1\_h/MatMul\_363]:256 \\ 
PACK & 13.822 & 0.628173\% & [bin\_dgcnn\_b\_f1\_h/MatMul\_364]:257 \\ 
SUB & 29.9335 & 1.3604\% & [bin\_dgcnn\_b\_f1\_h/sub\_3;bin\_dgcnn\_b\_f1\_h/MatMul\_3;b]:258 \\ 

  \end{tabular}
  \caption{LCE Profiler data for a single Hamming distance computation 
  as a matrix multiplication, in "BDGCNN BF H".}
  \label{tab:profiler2}
\end{table*}

These operations account for 24\% of the network's run time. Thus,
a speed-up of 32x of these operations would reduce them to around
1\% of the network's run time, which is negligible.

While we did not have an optimized version integrated with the LARQ runtime at 
the time of writing, optimizing the pairwise Hamming distance computation in binary space with
ARM NEON (SIMD) operations is quite simple, since it can be implemented
as $\text{popcount}(x \text{xor} y)$. On bit-packed 64-bit data (conversion
handled by LCE), with feature vectors of dimension 64, this can be written as:

\definecolor{backcolour}{rgb}{0.95,0.95,0.92}
\begin{lstlisting}[language=C, basicstyle=\scriptsize, captionpos=b, numbers=left, backgroundcolor=\color{backcolour}, tabsize=2, breaklines=true, 
caption=Implementation of pairwise Hamming distance in ARM NEON instrinsics (for readability).
Note that this code actually treats 64 features at a time\, and could thus provide a 64x speedup
(or more by grouping loads and writes with vld4). We use 32x as a conservative estimate since
we couldn't account for LCE's bit-packed conversion.
]
#include "arm_neon.h"

// input data in feats
int8_t n_outs = npoints*(npoints-1)/2
int8_t* out = malloc(n_outs*sizeof(int8_t));
for(int i = 0; i < npoints; i++) {

    // load first feature
    uint32x2_t a = vld1_u32(feats + 8*i);
    
    for(int j = i; j < npoints; j++) {
        
        //load second feature
        uint32x2_t b = vld1_u32(feats + 8*j);
        
        b = veor_u32(a, b); // XOR op
        
        // popcount op
        int8x8_t c = vreinterpret_u32_s8(b);
        c = vcnt_s8(c);
        
        // reduce to single number
        // by adding as a tree
        int64x1_t res;
        res = vpaddl_s32(vpaddl_s16(vpaddl_s8(c)));
        
        //store the output (last 8 bits)
        int8x8_t res8 = vreinterpret_s64_s8(res);
        out[j + npoints*j] = vget_lane_s8(res8, 7);
    }
}
\end{lstlisting}

"TopK" operations account for 22\% of the runtime and
we view them as incompressible in our simulation (Table \ref{tab:profiler}). 
It is possible that they could be written in NEON as well,
however, this optimization is not as trivial as the
Hamming distance one. Remaining operations, such as "Concatenation",
cannot be optimized further.

Contrary to simpler GNNs such as GCN, DGCNN is quite computationally intensive and involves a variety of operations on top of simple dot products, which makes it an interesting challenge for binarization, and illustrate that for complex graph neural networks more efforts are required, such as redefining suitable edge messages for binary graph features, or speeding-up pairwise distances computations, as done in this work. The inherent complexity also limits the attainable speedups from binarization, as shown by the large portion of the runtime taken by memory operations (concatenation) and top-k.

\section{Details regarding GraphSAGE}

In all experiments, the architecture used is identical to that used as a baseline by the OGB team. We report the accuracy following verbatim the experimental procedure of the OGB benchmark, using the suitable provided evaluators and dataset splits. Due to the very large number of edges in the dataset, we were unable to implement LSP in a sufficiently scalable manner (although the forward pass of the similarity computation can be implemented efficiently, the gradient of the similarity with respect to the node features is a tensor of size $|\mathcal{E}| \times |\mathcal{V}| \times D$ where $|\mathcal{E}|$ is the number of edges in the graph, $|\mathcal{V}|$ the number of nodes, and $D$ the dimension of the features. Although the tensor is sparse,  Pytorch currently did not have sufficient support of sparse tensors for gradients. We therefore chose not to include the results in the main text. We report the results of our binary GraphSAGE models, against two floating-point baselines: GraphSAGE and GCN.

\section{Balance functions}

For completeness, we also report the results at stage 2 of the multi-stage distillation scheme in Table \ref{tab:balance_stage2}. It is apparent that the additional operations degraded the performance not only for the full-binary models of stage 3, but also for the models for which all inputs are binary but weights are real.

\begin{table*}[b]
\centering
\begin{tabular}{c|c|c|c|c|c|c}
Model           & Stage & KNN   & LSP       & Global balance    &  Edge balance & Acc\\
\shline
BF2             & 2     & H     & -         & -                 &  Median       & 90.07\\
BF2             & 2     & H     & -         & -                 &  Mean         & 83.87\\
\hline
BF2             & 2     & H     & $\ell_2$  & Median            &               & 87.60\\
BF2             & 2     & H     & $\ell_2$  & Mean              &               & 89.47\\
\hline
Baseline BF2    & 2     & H     & $\ell_2$  & None              &               & \textbf{91.57}
\end{tabular}
\caption{Effect of additional balance functions on models with binary activations but floating-point weights. The performance of the baseline model suffers with the introduction of either mean or median centering prior to quantization.}
\label{tab:balance_stage2}
\end{table*}

\section{Table of mathematical operators}
\begin{table*}
    \centering
    \begin{tabular}{p{0.07\linewidth} | p{0.24\linewidth} | p{0.58\linewidth}}
         Symbol & Name & Description \\
         \shline
         $||.||_H$ & Hamming norm & Number of non-zero (or not -1) bits in a binary vector\\ \hline
         $d(., .)_H$ & Hamming distance & Number of bits that differ between two binary vectors, equivalent to \texttt{popcount(xor())}\\ \hline
         $\oplus$ & Exclusive OR (XOR) & $1 \oplus 1 = -1 \oplus -1 = -1, \quad -1 \oplus 1 = 1 \oplus -1 = 1$\\ \hline
         $\odot$ & Hadamard product & Element-wise product between tensors\\ \hline
         $\circledast$ & Binary-real or Binary-binary dot product or convolution & Equivalent to \texttt{popcount(xnor())} (\ie no multiplications) for binary tensors\\ \hline
         $\otimes$ & Outer product & \\ \hline
         $\star$ & Dot product or convolution & Denoted by $\ast$ in \cite{Rastegari2016} \\ \hline
         $|\mathcal{X}|$ & Cardinal of a set $\mathcal{X}$ & Number of elements in the set\\ \hline
         $\mathbf{x}^{(l)}$ & Feature maps at layer $l$ & \\ \hline
         $. || .$ & Concatenation &\\ \hline
         $\coloneqq$ & Definition &\\ \hline
         $\mathbf{x}^{(l)}$ & Element $\mathbf{x}$ at layer $l$\\ \hline
    \end{tabular}
    \caption{Table of the mathematical operators used in the manuscript.}
    \label{tab:mathoperators}
\end{table*}

{\small
\bibliographystyle{ieee_fullname}
\bibliography{bingnn}

\begin{thebibliography}{10}\itemsep=-1pt

\bibitem{Alon2020}
Uri Alon and Eran Yahav.
\newblock On the bottleneck of graph neural networks and its practical
  implications.
\newblock In {\em ICLR}, 2021.

\bibitem{bahl2019low}
Ga{\'{e}}tan Bahl, Lionel Daniel, Matthieu Moretti, and Florent Lafarge.
\newblock {Low-power neural networks for semantic segmentation of satellite
  images}.
\newblock In {\em ICCV Workshops}, 2019.

\bibitem{bengio2013estimating}
Yoshua Bengio, Nicholas L{\'e}onard, and Aaron Courville.
\newblock Estimating or propagating gradients through stochastic neurons for
  conditional computation.
\newblock {\em arXiv:1308.3432}, 2013.

\bibitem{bronstein_numerical_2008}
Alexander Bronstein, Michael Bronstein, and Ron Kimmel.
\newblock {\em {Numerical Geometry of Non-Rigid Shapes}}.
\newblock Springer Publishing Company, Incorporated, 1 edition, 2008.

\bibitem{Bronstein2017}
Michael~M Bronstein, Joan Bruna, Yann LeCun, Arthur Szlam, and Pierre
  Vandergheynst.
\newblock Geometric deep learning: going beyond euclidean data.
\newblock {\em IEEE Signal Processing Magazine}, 34(4), 2017.

\bibitem{Bruna2013b}
Joan Bruna, Wojciech Zaremba, Arthur Szlam, and Yann LeCun.
\newblock {Spectral networks and deep locally connected networks on graphs}.
\newblock {\em ICLR}, 2014.

\bibitem{DBLP:conf/bmvc/BulatT19}
Adrian Bulat and Georgios Tzimiropoulos.
\newblock {XNOR-Net++: Improved binary neural networks}.
\newblock In {\em BMVC}, 2019.

\bibitem{Bulat2019}
Adrian Bulat, Georgios Tzimiropoulos, Jean Kossaifi, and Maja Pantic.
\newblock {Improved training of binary networks for human pose estimation and
  image recognition}.
\newblock {\em arXiv:1904.05868}, 2019.

\bibitem{Casas2020SpAGNNSG}
Sergio Casas, Cole Gulino, Renjie Liao, and R. Urtasun.
\newblock Spagnn: Spatially-aware graph neural networks for relational behavior
  forecasting from sensor data.
\newblock {\em International Conference on Robotics and Automation (ICRA)},
  2020.

\bibitem{Qi2017a}
R~Qi Charles, Hao Su, Mo Kaichun, and Leonidas~J Guibas.
\newblock {PointNet: Deep Learning on Point Sets for 3D Classification and
  Segmentation}.
\newblock In {\em CVPR}, 2017.

\bibitem{Chiang2019}
Wei~Lin Chiang, Yang Li, Xuanqing Liu, Samy Bengio, Si Si, and Cho~Jui Hsieh.
\newblock {Cluster-GCN: An efficient algorithm for training deep and large
  graph convolutional networks}.
\newblock In {\em Proceedings of the ACM SIGKDD International Conference on
  Knowledge Discovery and Data Mining}, 2019.

\bibitem{defferrard_convolutional_2016}
Micha\"{e}l Defferrard, Xavier Bresson, and Pierre Vandergheynst.
\newblock Convolutional neural networks on graphs with fast localized spectral
  filtering.
\newblock In D.~D. Lee, M. Sugiyama, U.~V. Luxburg, I. Guyon, and R. Garnett,
  editors, {\em NeurIPS 29}. 2016.

\bibitem{derrible2011applications}
Sybil Derrible and Christopher Kennedy.
\newblock {Applications of graph theory and network science to transit network
  design}.
\newblock {\em Transport reviews}, 31(4), 2011.

\bibitem{Fey/Lenssen/2019}
Matthias Fey and Jan~E Lenssen.
\newblock {Fast Graph Representation Learning with PyTorch Geometric}.
\newblock In {\em ICLR Workshop on Representation Learning on Graphs and
  Manifolds}, 2019.

\bibitem{Fey2017}
Matthias Fey, Jan~Eric Lenssen, Frank Weichert, and Heinrich Muller.
\newblock {SplineCNN: Fast Geometric Deep Learning with Continuous B-Spline
  Kernels}.
\newblock {\em CVPR}, 2018.

\bibitem{Frasca2020}
Fabrizio Frasca, Emanuele Rossi, Davide Eynard, Benjamin Chamberlain, Michael
  Bronstein, and Federico Monti.
\newblock Sign: Scalable inception graph neural networks.
\newblock In {\em ICML 2020 Workshop on Graph Representation Learning and
  Beyond}, 2020.

\bibitem{Geiger2020}
Lukas Geiger and Plumerai Team.
\newblock {Larq: An Open-Source Library for Training Binarized Neural
  Networks}.
\newblock {\em Journal of Open Source Software}, 5(45), 2020.

\bibitem{Gilmer2017}
Justin Gilmer, Samuel~S. Schoenholz, Patrick~F. Riley, Oriol Vinyals, and
  George~E. Dahl.
\newblock {Neural message passing for quantum chemistry}.
\newblock In {\em ICML}, 2017.

\bibitem{Gong2020}
Shunwang Gong, Mehdi Bahri, Michael~M. Bronstein, and Stefanos Zafeiriou.
\newblock {Geometrically Principled Connections in Graph Neural Networks}.
\newblock In {\em CVPR}, 2020.

\bibitem{gong2014compressing}
Yunchao Gong, Liu Liu, Ming Yang, and Lubomir Bourdev.
\newblock Compressing deep convolutional networks using vector quantization.
\newblock {\em arXiv:1412.6115}, 2014.

\bibitem{Gori2005}
Marco Gori, Gabriele Monfardini, and Franco Scarselli.
\newblock {A new model for learning in Graph domains}.
\newblock {\em International Joint Conference on Neural Networks}, 2(May 2014),
  2005.

\bibitem{Hamilton2017}
William~L. Hamilton, Rex Ying, and Jure Leskovec.
\newblock {Inductive representation learning on large graphs}.
\newblock In {\em NeurIPS}, 2017.

\bibitem{44873}
Geoffrey Hinton, Oriol Vinyals, and Jeffrey Dean.
\newblock {Distilling the Knowledge in a Neural Network}.
\newblock In {\em NIPS Deep Learning and Representation Learning Workshop},
  2015.

\bibitem{hu2020open}
Weihua Hu, Matthias Fey, Marinka Zitnik, Yuxiao Dong, Hongyu Ren, Bowen Liu,
  Michele Catasta, and Jure Leskovec.
\newblock Open graph benchmark: Datasets for machine learning on graphs.
\newblock In {\em NeurIPS}, 2020.

\bibitem{huang2020skipgnn}
Kexin Huang, Cao Xiao, Lucas~M Glass, Marinka Zitnik, and Jimeng Sun.
\newblock Skipgnn: predicting molecular interactions with skip-graph networks.
\newblock {\em Scientific reports}, 10(1), 2020.

\bibitem{NIPS2016_d8330f85}
Itay Hubara, Matthieu Courbariaux, Daniel Soudry, Ran El-Yaniv, and Yoshua
  Bengio.
\newblock {Binarized Neural Networks}.
\newblock In D Lee, M Sugiyama, U Luxburg, I Guyon, and R Garnett, editors,
  {\em NeurIPS}, 2016.

\bibitem{Kazi2020}
Anees Kazi, Luca Cosmo, Nassir Navab, and Michael Bronstein.
\newblock Differentiable graph module (dgm) for graph convolutional networks.
\newblock {\em arXiv:2002.04999}, 2020.

\bibitem{khanfor2020graph}
Abdullah Khanfor, Amal Nammouchi, Hakim Ghazzai, Ye Yang, Mohammad~R Haider,
  and Yehia Massoud.
\newblock Graph neural networks-based clustering for social internet of things.
\newblock In {\em International Midwest Symposium on Circuits and Systems
  (MWSCAS)}, 2020.

\bibitem{Kingma2014}
Diederik~P Kingma and Jimmy Ba.
\newblock Adam: A method for stochastic optimization.
\newblock {\em arXiv:1412.6980}, 2014.

\bibitem{Kipf2017}
Thomas~N. Kipf and Max Welling.
\newblock {Semi-supervised classification with graph convolutional networks}.
\newblock In {\em ICLR}, 2017.

\bibitem{8463534}
H Lai, Y Pan, S Liu, Z Weng, and J Yin.
\newblock {Improved Search in Hamming Space Using Deep Multi-Index Hashing}.
\newblock {\em IEEE T-PAMI}, 29(9), 2019.

\bibitem{Lee2019}
Seunghyun Lee and Byung~Cheol Song.
\newblock Graph-based knowledge distillation by multi-head attention network.
\newblock In {\em BMVC}, 2019.

\bibitem{lemaire2020fpga}
Edgar Lemaire, Matthieu Moretti, Lionel Daniel,
  Beno$\backslash${\^{}}$\backslash$it Miramond, Philippe Millet, Frederic
  Feresin, and S{\'{e}}bastien Bilavarn.
\newblock {An FPGA-based Hybrid Neural Network accelerator for embedded
  satellite image classification}.
\newblock In {\em IEEE International Symposium on Circuits and Systems
  (ISCAS)}, 2020.

\bibitem{Li2019}
Guohao Li, Matthias Muller, Ali Thabet, and Bernard Ghanem.
\newblock {DeepGCNs: Can GCNs go as deep as CNNs?}
\newblock In {\em ICCV}, 2019.

\bibitem{Liu2012}
Wei Liu, Jun Wang, Rongrong Ji, Yu~Gang Jiang, and Shih~Fu Chang.
\newblock {Supervised hashing with kernels}.
\newblock In {\em CVPR}, 2012.

\bibitem{zitnik2018biosnap}
Sagar~Maheshwari Marinka~Zitnik, Rok~Sosi\v{c} and Jure Leskovec.
\newblock Biosnap datasets: Stanford biomedical network dataset collection.
\newblock \url{http://snap.stanford.edu/biodata}, Aug. 2018.

\bibitem{real2binICLR20}
Brais Martinez, Jing Yang, Adrian Bulat, and Georgios Tzimiropoulos.
\newblock {Training binary neural networks with real-to-binary convolutions}.
\newblock In {\em ICLR}, 2020.

\bibitem{mirza2003studying}
Batul~J Mirza, Benjamin~J Keller, and Naren Ramakrishnan.
\newblock {Studying recommendation algorithms by graph analysis}.
\newblock {\em Journal of intelligent information systems}, 20(2), 2003.

\bibitem{Monti2017}
Federico Monti, Davide Boscaini, Jonathan Masci, Emanuele Rodol{\'{a}}, Jan
  Svoboda, and Michael~M Bronstein.
\newblock {Geometric deep learning on graphs and manifolds using mixture model
  CNNs}.
\newblock {\em CVPR}, 2017.

\bibitem{motl2015ctu}
Jan Motl and Oliver Schulte.
\newblock The ctu prague relational learning repository.
\newblock {\em arXiv:1511.03086}, 2015.

\bibitem{10.3389/fgene.2019.00381}
Walter Nelson, Marinka Zitnik, Bo Wang, Jure Leskovec, Anna Goldenberg, and
  Roded Sharan.
\newblock {To Embed or Not: Network Embedding as a Paradigm in Computational
  Biology}.
\newblock {\em Frontiers in Genetics}, 10, 2019.

\bibitem{Norouzi2011}
Mohammad Norouzi and David~J. Fleet.
\newblock {Minimal loss hashing for compact binary codes}.
\newblock In {\em ICML}, 2011.

\bibitem{Norouzi2012}
Mohammad Norouzi, David~J. Fleet, and Ruslan Salakhutdinov.
\newblock {Hamming distance metric learning}.
\newblock In {\em NeurIPS}, 2012.

\bibitem{Norouzi2014}
Mohammad Norouzi, Ali Punjani, and David~J. Fleet.
\newblock {Fast exact search in hamming space with multi-index hashing}.
\newblock {\em IEEE T-PAMI}, 2014.

\bibitem{Park2019}
Wonpyo Park, Dongju Kim, Yan Lu, and Minsu Cho.
\newblock {Relational knowledge distillation}.
\newblock In {\em CVPR}, 2019.

\bibitem{NEURIPS2019_9015}
Adam Paszke, Sam Gross, Francisco Massa, Adam Lerer, James Bradbury, Gregory
  Chanan, Trevor Killeen, Zeming Lin, Natalia Gimelshein, Luca Antiga, Alban
  Desmaison, Andreas Kopf, Edward Yang, Zachary DeVito, Martin Raison, Alykhan
  Tejani, Sasank Chilamkurthy, Benoit Steiner, Lu Fang, Junjie Bai, and Soumith
  Chintala.
\newblock Pytorch: An imperative style, high-performance deep learning library.
\newblock In H. Wallach, H. Larochelle, A. Beygelzimer, F. d\textquotesingle
  Alch\'{e}-Buc, E. Fox, and R. Garnett, editors, {\em NeurIPS}. 2019.

\bibitem{Qin2020}
Haotong Qin, Zhongang Cai, Mingyuan Zhang, Yifu Ding, Haiyu Zhao, Shuai Yi,
  Xianglong Liu, and Hao Su.
\newblock Bipointnet: Binary neural network for point clouds.
\newblock In {\em ICLR}, 2021.

\bibitem{Rastegari2016}
Mohammad Rastegari, Vicente Ordonez, Joseph Redmon, and Ali Farhadi.
\newblock {XNOR-Net : ImageNet Classification Using Binary}.
\newblock {\em ECCV}, 2016.

\bibitem{Scarselli2009}
Franco Scarselli, Marco Gori, Ah~Chung Tsoi, Markus Hagenbuchner, and Gabriele
  Monfardini.
\newblock {The graph neural network model.}
\newblock {\em IEEE Trans. on Neural Networks}, 20(1), 2009.

\bibitem{Tung2019}
Frederick Tung and Greg Mori.
\newblock Similarity-preserving knowledge distillation.
\newblock In {\em ICCV}, 2019.

\bibitem{Velickovic2017}
Petar Veli{\v{c}}kovi{\'{c}}, Guillem Cucurull, Arantxa Casanova, Adriana
  Romero, Pietro Li{\'{o}}, and Yoshua Bengio.
\newblock {Graph Attention Networks}.
\newblock {\em ICLR}, 2018.

\bibitem{Verma2018}
Nitika Verma, Edmond Boyer, and Jakob Verbeek.
\newblock {FeaStNet: Feature-Steered Graph Convolutions for 3D Shape Analysis}.
\newblock In {\em CVPR}, 2018.

\bibitem{Wang2020}
Hanchen Wang, Defu Lian, Ying Zhang, Lu Qin, Xiangjian He, Yiguang Lin, and
  Xuemin Lin.
\newblock Binarized graph neural network.
\newblock {\em arXiv:2004.11147}, 2020.

\bibitem{Wang2020_bigcn}
Junfu Wang, Yunhong Wang, Zhen Yang, Liang Yang, and Yuanfang Guo.
\newblock Bi-gcn: Binary graph convolutional network.
\newblock {\em arXiv:2010.07565}, 2020.

\bibitem{Wang2019}
Yue Wang, Yongbin Sun, Ziwei Liu, Sanjay~E. Sarma, Michael~M. Bronstein, and
  Justin~M. Solomon.
\newblock {Dynamic graph Cnn for learning on point clouds}.
\newblock {\em ACM Trans. on Graphics}, 2019.

\bibitem{Wang2018_dgcnn}
Yue Wang, Yongbin Sun, Ziwei Liu, Sanjay~E Sarma, Michael~M Bronstein, and
  Justin~M Solomon.
\newblock {Dynamic Graph CNN for Learning on Point Clouds}.
\newblock {\em ACM Trans. on Graphics}, 38(5), 2019.

\bibitem{Wu2019}
Felix Wu, Tianyi Zhang, Amauri~Holanda de Souza, Christopher Fifty, Tao Yu, and
  Kilian~Q. Weinberger.
\newblock {Simplifying graph convolutional networks}.
\newblock In {\em ICML}, 2019.

\bibitem{Wu2020}
Zonghan Wu, Shirui Pan, Fengwen Chen, Guodong Long, Chengqi Zhang, and
  Philip~S. Yu.
\newblock {A Comprehensive Survey on Graph Neural Networks}.
\newblock {\em IEEE Trans. on Neural Networks and Learning Systems}, 2020.

\bibitem{Wu2015}
Zhirong Wu, Shuran Song, Aditya Khosla, Fisher Yu, Linguang Zhang, Xiaoou Tang,
  and Jianxiong Xiao.
\newblock {3D ShapeNets: A deep representation for volumetric shapes}.
\newblock In {\em CVPR}, 2015.

\bibitem{Yang_2020_CVPR}
Yiding Yang, Jiayan Qiu, Mingli Song, Dacheng Tao, and Xinchao Wang.
\newblock {Distilling Knowledge From Graph Convolutional Networks}.
\newblock In {\em CVPR}, 2020.

\bibitem{Zagoruyko2017AT}
Sergey Zagoruyko and Nikos Komodakis.
\newblock {Paying More Attention to Attention: Improving the Performance of
  Convolutional Neural Networks via Attention Transfer}.
\newblock In {\em ICLR}, 2017.

\bibitem{graphsaint-iclr20}
Hanqing Zeng, Hongkuan Zhou, Ajitesh Srivastava, Rajgopal Kannan, and Viktor
  Prasanna.
\newblock {GraphSAINT: Graph Sampling Based Inductive Learning Method}.
\newblock In {\em ICLR}, 2020.

\bibitem{zhangequipment}
Weishan Zhang, Yafei Zhang, Liang Xu, Jiehan Zhou, Yan Liu, Mu Gu, Xin Liu, and
  Su Yang.
\newblock Modeling iot equipment with graph neural networks.
\newblock {\em IEEE Access}, PP, 2019.

\bibitem{zhou2016dorefa}
Shuchang Zhou, Yuxin Wu, Zekun Ni, Xinyu Zhou, He Wen, and Yuheng Zou.
\newblock Dorefa-net: Training low bitwidth convolutional neural networks with
  low bitwidth gradients.
\newblock {\em arXiv:1606.06160}, 2016.

\end{thebibliography}
}

\end{document}